\documentclass[11pt]{article}

\usepackage[preprint]{acl}

\usepackage{times}
\usepackage{latexsym}

\usepackage[T1]{fontenc}

\usepackage[utf8]{inputenc}

\usepackage{microtype}

\usepackage{inconsolata}

\usepackage{graphicx}

\usepackage[acronym]{glossaries}
\glsdisablehyper

\usepackage[table,xcdraw]{xcolor}
\usepackage{pifont}
\usepackage{tikz}
\usepackage{multirow}
\usepackage{xurl}
\newcommand*\emptycircled[1]{\tikz[baseline=(char.base)]{
            \node[shape=circle,draw,inner sep=.4pt] (char) {#1};}}

\title{Wyvern: An Agentic Framework for Generating Grounded Multimodal Reports}

\author{
 \textbf{Beatrice Alessandra Motetti\textsuperscript{1,2}
 },
 \textbf{Emilien Guandalino\textsuperscript{2}},
 \textbf{Daniele Jahier Pagliari\textsuperscript{1}},
 \\
 \textbf{Alessio Burrello\textsuperscript{1}},
 \textbf{Lorenz K. Müller\textsuperscript{2}},
 \textbf{Konstantin Berestizshevsky\textsuperscript{2}},
 \textbf{Lukas Cavigelli\textsuperscript{2}}
\\
 \textsuperscript{1}Politecnico di Torino, Italy
 \textsuperscript{2}Computing Systems Lab,  Huawei Research, Switzerland
\\
 \small{
   \textbf{Correspondence:} \href{mailto:beatrice.motetti@polito.it}{\texttt{beatrice.motetti@polito.it}}
 }
}

\begin{document}
\maketitle
\begin{abstract}
In the current artificial intelligence-driven innovation era, the pace of knowledge growth is accelerating, and is hard to keep up with. While generative models are increasingly used to synthesize content, they often lack in information grounding.
To address these peculiarities of our time, we propose Wyvern, a multi-agent framework for the automated generation of grounded, multimodal technical reports. 
Wyvern allows for the generation of multimodal outputs, integrating images, tables, and text with supporting references in a unified report. Additionally, a particular focus is placed on the grounding of the content, with the implementation of a claims auto-revision stage. 
We conduct a human evaluation study to assess the quality of our proposed framework. The results show that the figures' informativeness is perceived as superior to that of a recent baseline in 87\% of cases. Furthermore, Wyvern's reports are rated as more useful than those produced by three alternative methods in 63\% to 100\% of instances. 
We also carry out automatic evaluations showing that Wyvern gains up to 2.3$\times$ in citation recall and 1.6$\times$ in citation precision with respect to the baselines.

\end{abstract}
\newacronym{llm}{LLM}{Large Language Model}
\newacronym{rag}{RAG}{Retrieval-Augmented Generation}
\newacronym{nli}{NLI}{Natural Language Inference}
\newacronym{lrm}{LRM}{Large Reasoning Model}
\newacronym{vlm}{VLM}{Vision Language Model}
\newacronym{cot}{CoT}{Chain-of-Thought}

\section{Introduction}
\label{sec:introduction}

The volume of newly published content, spanning from scientific literature to general webpages, is rapidly growing, particularly in fields such  as artificial intelligence~\cite{bornmann_growth_2021}. Keeping up with the vast quantity of heterogeneous domain-specific information has become extremely critical and complex simultaneously, especially for researchers~\cite{park_papers_2023, jones_burden_2009}. 
In this context, automated synthesis and summarization tools can act as first-pass filters that retrieve, aggregate and condense large and relevant information into technical reports, helping researchers and end-users to stay informed and keep up in fast-evolving fields. While these tools cannot substitute critical human judgment, they can ease the burden to efficiently retrieve and manage information.

\begin{figure}[]
  \centering
  \includegraphics[width=\linewidth]{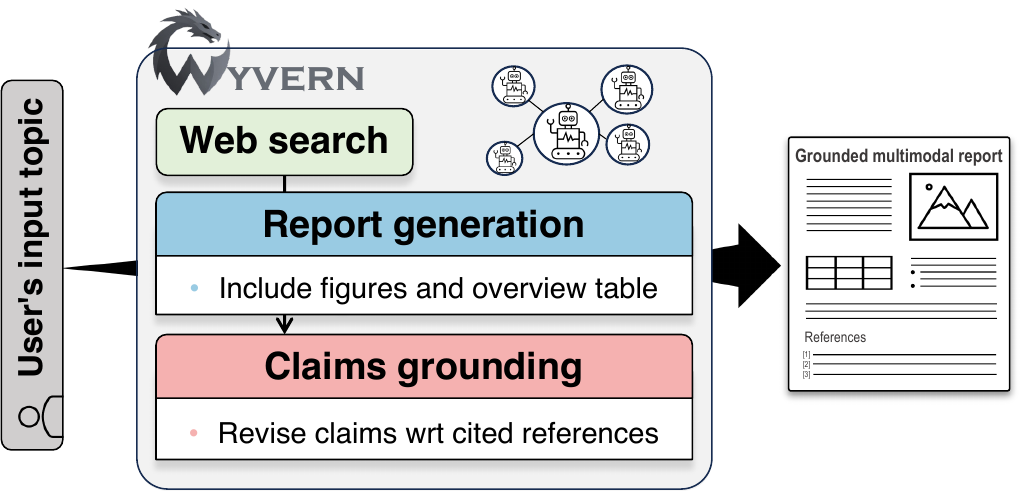}
  \caption{Overview of the main functionalities of Wyvern}
  \label{fig:overview}
\end{figure}

Several frameworks have been proposed recently to automate knowledge retrieval and its organization into machine-generated textual documents~\cite{shao2024storm, wang2024autosurvey, jiang2024costorm}. Furthermore, as multimodality allows for more engaging content~\cite{fu_doc2ppt_2022}, recent works have also considered the inclusion of relevant images in the output reports to increase their informativeness~\cite{yang2025wikiautogen, yang2025multimodaldeepresearchergeneratingtextchart}.
A key requirement for these frameworks is the ability to ground generated statements on retrieved evidence, thereby increasing user trust. Traceability between claims and sources is fundamental, as it enables to explore with more depth the aspects of interest.

Building on these observations, we propose Wyvern, a multi-agent framework for the automatic generation of multimodal technical reports on user-selected topics, with a particular focus on content groundedness. Our Wyvern framework capitalizes on a structured workflow of specialized \gls{llm}-based agents, each focused on a narrow task, such as refining the quality of previously acquired information.
Organizing these agents into a coordinated framework significantly enhances reliability, modularity, and overall quality of the produced content, as recent works have demonstrated~\cite{huang2024agentcodermultiagentbasedcodegeneration,mas_software}.  As illustrated in Figure~\ref{fig:overview}, Wyvern comprises various modules that search and process relevant information from the web, produce the multimodal report, and verify whether the textual statements are supported by the previously retrieved evidence.

We evaluate Wyvern through a human evaluation study, comparing its generated reports with those produced by STORM~\cite{shao2024storm}, WebThinker~\cite{Li2025WebThinker} and WikiAutoGen~\cite{yang2025wikiautogen}. The results show that Wyvern produces reports whose figures and factuality are more positively evaluated by the end-users. We also conduct an automatic evaluation, and examine the predictive power of different evaluation means with respect to human judgment.

Our key contributions can be summarized as follows:
\begin{itemize}
    \item We propose Wyvern, a multi-agent framework for the generation of grounded, multimodal technical reports, retrieving the most relevant information from the web. As key novelties:
    \begin{itemize}
        \item we design a module for the retrieval, selection, and positioning of the most informative images within the textual report;
        \item we build a claims auto-revision stage to verify the grounding of the statements on the collected evidence.
    \end{itemize}
    \item We assess the quality of Wyvern-generated reports through a human evaluation study and we compare these results with automatic evaluations, to analyze the predictive performance of the latter for human judgments.
\end{itemize}

Code and data are available at \url{https://github.com/huawei-csl/wyvern}.

\section{Related Work}
\label{sec:related_works}

\subsection{Automated long-form expository writing}

The generation of long text by \glspl{llm} is a complex task, due to their limited context length and the need to maintain semantic consistency~\cite{wang2024autosurvey, yang2025wikiautogen}. To overcome the limitations of static parametric knowledge, a common approach is to integrate \gls{rag} strategies~\cite{lewis_rag_2020}, often combined with web search engines~\cite{shao2024storm, Li2025WebThinker}, to perform an up-to-date and comprehensive information retrieval. 

Many recent works on long-form expository writing adopt a top-down approach, with an initial outline definition and a subsequent section expansion phase~\cite{shao2024storm, jiang2024costorm, yang2025wikiautogen, wang2024autosurvey, yang2025multimodaldeepresearchergeneratingtextchart}; others employ a more dynamic planning strategy that builds upon recursive task decomposition principles~\cite{Li2025WebThinker,xiong2025writehere}.
Specifically, STORM~\cite{shao2024storm} and Co-STORM~\cite{jiang2024costorm} employ ensembles of multi-perspective agents with a question-and-answer approach for information retrieval and outline planning tasks, generating Wikipedia-like articles. AutoSurvey~\cite{wang2024autosurvey} tackles the problem of creating literature surveys with a four-stage framework comprising information retrieval and outline definition, and subsections drafting, integration, and evaluation. WebThinker~\cite{Li2025WebThinker} integrates the capabilities of performing web searches and report writing and editing within the reasoning of Large Reasoning Models.

Another emerging line of works tackles the generation of multimodal reports, incorporating retrieved images within the textual body~\cite{yang2025wikiautogen, yang2025multimodaldeepresearchergeneratingtextchart}. In WikiAutoGen~\cite{yang2025wikiautogen} an agent proposes candidate positions and descriptions of images, then retrieved from the web.
Multimodal DeepResearcher~\cite{yang2025multimodaldeepresearchergeneratingtextchart} uses instead a structured textual representation of charts to allow for the direct generation of visualizations, by using a \gls{llm} to produce JavaScript code.

We report in Table~\ref{tab:related_comparison} an overview of the features of the main works.

\begin{table}[]
\centering
\setlength{\tabcolsep}{1mm}
\resizebox{\columnwidth}{!}{%
\begin{tabular}{ccc}
\hline
\textbf{Method} & \textbf{\begin{tabular}[c]{@{}c@{}}Images \\ insertion\end{tabular}} & \textbf{\begin{tabular}[c]{@{}c@{}}Claims\\ verification\end{tabular}} \\ \hline
\rowcolor[HTML]{EFEFEF} 
STORM~\cite{shao2024storm}                                                                & \ding{55}                        & \ding{55}                        \\
WebThinker~\cite{Li2025WebThinker}                                                           & \ding{55}                        & \ding{55}                        \\
\rowcolor[HTML]{EFEFEF} 
WikiAutoGen~\cite{yang2025wikiautogen}                                                          & \ding{51}                        & \ding{55}                        \\

{\color[HTML]{000000} \textbf{Wyvern (ours)}}                                    & {\color[HTML]{000000} \ding{51}} & {\color[HTML]{000000} \ding{51}} \\
\hline
\end{tabular}%
}
\caption{Overview of expository text generators. Wyvern integrates multimodality and explicit claims verification and revision into a unified framework.}
\label{tab:related_comparison}
\end{table}
\subsection{Grounding}
Measuring and improving the factuality of text generated by \glspl{llm} is an active area of research, with two main directions~\cite{jacovi2025factsgroundingleaderboardbenchmarking}: one defines factuality as text entailment with respect to cited sources; the other verifies claims against external ground-truth answers. 
Additionally, various definitions of claim unit exist, with some works identifying it at a sentence level~\cite{gao_enabling_2023, shao2024storm, balepur2023imitate}, and others at a finer granularity by decomposing sentences into atomic facts~\cite{min-etal-2023-factscore, li-etal-2024-self}. 

\begin{figure*}[]
  \centering
  \includegraphics[width=\linewidth]{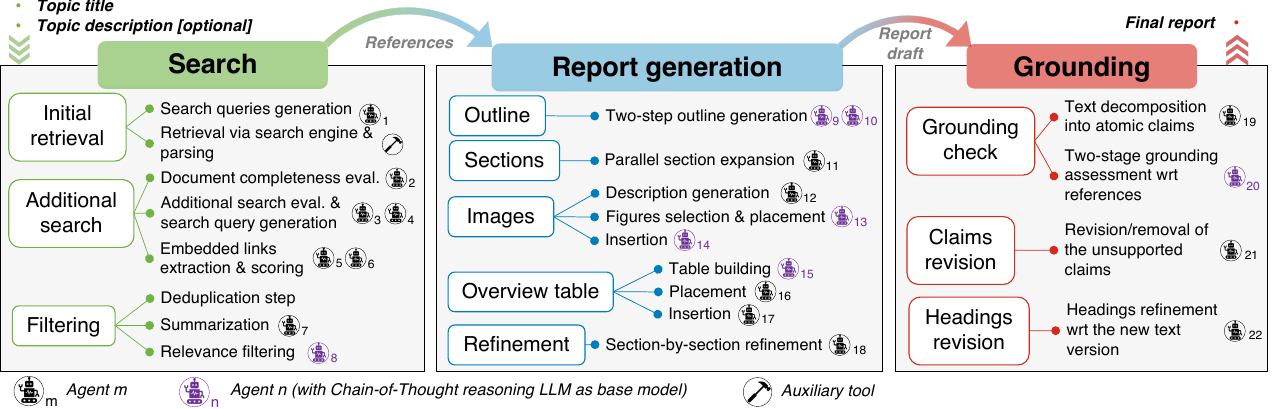}
  \caption{Overview of the key components of Wyvern, our proposed framework. The search module (left) retrieves relevant documents from the web. The report generation module (middle) handles of the writing of the report's text and of the figures insertion. The grounding module (right) verifies if the claims are supported by the cited references, and possibly revises them.
  }
  \label{fig:method}
\end{figure*}

FActSCORE~\cite{min-etal-2023-factscore} decomposes sentences into atomic facts, to overcome the challenge of assigning a binary entailment label to sentences with multiple pieces of information.
Self-Checker~\cite{li-etal-2024-self} first decomposes the text into claims, and then generates queries to retrieve relevant content to verify them. Chain-of-Verification~\cite{dhuliawala-etal-2024-chain} mitigates hallucinations in baseline \glspl{llm} responses by generating a list of verification questions for the key claims, and then revising the responses via self-validation using the model’s parametric knowledge.

\citet{gao_enabling_2023} measure groundedness with respect to the cited sources by computing the citation recall and precision over sentence-level statements.
Differently, we define the citation quality metrics on atomic claims derived from the full-sentences, avoiding cases of \textit{partial} support.

\section{Proposed framework}
\label{sec:proposed_framework}

Wyvern comprises distinct modules that operate sequentially on individual macro-tasks (see Figure~\ref{fig:method}). Each of the modules handles a specific phase of the report generation by relying on ensembles of agents. In particular, Wyvern integrates: (i) a \textit{search} module, that is tasked with the retrieval and processing of relevant information; (ii) a \textit{report generation} module, that writes the report and integrates it with the most informative figures from the retrieved documents, and with an overview table; (iii) a \textit{grounding} module, which examines all the statements of the report with the aim of improving its factuality and limiting the hallucinations and the unsupported statements.
In the following, we refer to each agent involved in our framework by the subscript number next to its icon in Figure~\ref{fig:method}.

\subsection{Search phase}
The goal of the search phase is to acquire information from the web and construct a \textit{knowledge database} consisting in a set of references $R$ and their corresponding content summaries $R_S$. The search begins with the user providing Wyvern with a topic title, optionally specifying aspects of interest or non-interest and a list of relevant URLs. Afterwards a \gls{llm} generates a refined topic title and a detailed description of it, and the URLs are directly incorporated into the initial references base~$R$. 

Wyvern's agent~\emptycircled{1} then leverages the topic title and description to generate $n$ search queries. For each search query, it retrieves the top-$k_1$ URLs from a search engine. The type of webpage content determines the selection of the most suitable parser, which is used to extract the URL's textual content into Markdown format. To assure content quality and processability, only documents whose length falls within a predefined token range are retained. This removes both empty pages and documents that are too large for the \gls{llm}'s context length. 

To refine the set of documents to be included in the knowledge database, Wyvern employs an ensemble of independent agents that analyze the retrieved information. Agent~\emptycircled{2} assigns a score on a 10-point scale to assess the completeness of the extracted Markdown content for each webpage, in parallel. Documents with a completeness score below a predefined threshold $\lambda_\text{comp}$ are excluded from the references base, thereby preserving content quality by removing the documents for which the parser failed to extract meaningful text.
Agent~\emptycircled{3} evaluates, for each webpage, the need for an additional search step to retrieve additional relevant material. Similarly to agent~\emptycircled{2}, it assigns a score on a 10-point scale to each document, estimating whether additional informative content can be fetched with a new search. If the score exceeds a set threshold $\lambda_\text{search}$, agent~\emptycircled{4} generates a candidate search query, which is used to fetch the most relevant URL. This URL is then considered for inclusion in the reference base, following the parsing and completeness evaluation procedure carried out by agent~\emptycircled{2}.

To broaden the coverage of relevant aspects, Wyvern analyzes the embedded links within the parsed documents. This is done in a two-stage process. First, the embedded hyperlinks are extracted from each collected resource by agent~\emptycircled{5}, and the relative links are resolved to their absolute versions. Afterwards, for each document, agent~\emptycircled{6} is provided with the textual content, the list of extracted hyperlinks, and the user-provided topic of interest, and is tasked with assigning a relevance score to each link. Links scoring below a set threshold $\lambda_\text{link}$ are considered out of scope, while the top-$k_2$ by score are processed, following the same content extraction process as the initial documents. This iterative exploration process enables Wyvern to dynamically enrich the collected knowledge base.

To minimize redundancy in the reference base, a deduplication step is carried out to remove the most similar documents (e.g., earlier drafts or alternative versions of the same paper or webpage). Specifically, the Jaccard similarity index between all pairs of documents is computed. For each pair exceeding a similarity threshold $\lambda_\text{sim}$,  only the longer document is kept in the final reference base $R$.

Once the reference base is finalized, agent~\emptycircled{7} summarizes each document $d_i \in R$, thus creating the set of reference summaries $R_S$ that will be used for report generation. Eventually, agent~\emptycircled{8}, based on a Chain-of-Thought reasoning \gls{llm}, conducts a final filtering step to detect semantic outliers in $R_S$, i.e. documents that are off-topic or partially misaligned with the original topic description. The agent considers all the summaries of the references simultaneously, to infer the underlying domain, overcoming the limitation of relying only on the \gls{llm}'s static knowledge about the input topic. In this way, documents which are not aligned to the topic can be flagged as outliers and excluded from both $R_S$ and $R$, allowing for more focused content.

\subsection{Report generation}

\subsubsection{Outline creation and sections expansion}
Similarly to prior works~\cite{shao2024storm, wang2024autosurvey}, the generation of the textual content of the report follows a two-stage process, comprising outline creation and section expansion. 
The adopted approach is inspired by the exploration-exploitation trade-off. 
The set of summarized references $R_S$ is used by agent~\emptycircled{9} to generate $N_o$ draft outlines $\{O_1, ..., O_{N_o}\}$ to maximize topic coverage. Agent~\emptycircled{10} then merges the generated outline versions to produce a final outline~$O$ including the most informative aspects of the explored options.

Once the outline $O$ is produced, agent~\emptycircled{11} generates the text of each section $s \in O$ simultaneously, with full access to the references summaries set~$R_S$. For each first-level heading, the agent writes the section and its associated subsections, following the outline $O$. The agent is thus responsible for selecting the most salient content and placing citations to the supporting references. Eventually, the concatenation of all the expanded sections is performed to obtain the draft of the report.

\subsubsection{Image insertion}

\begin{figure}[]
  \centering
  \includegraphics[width=1\linewidth]{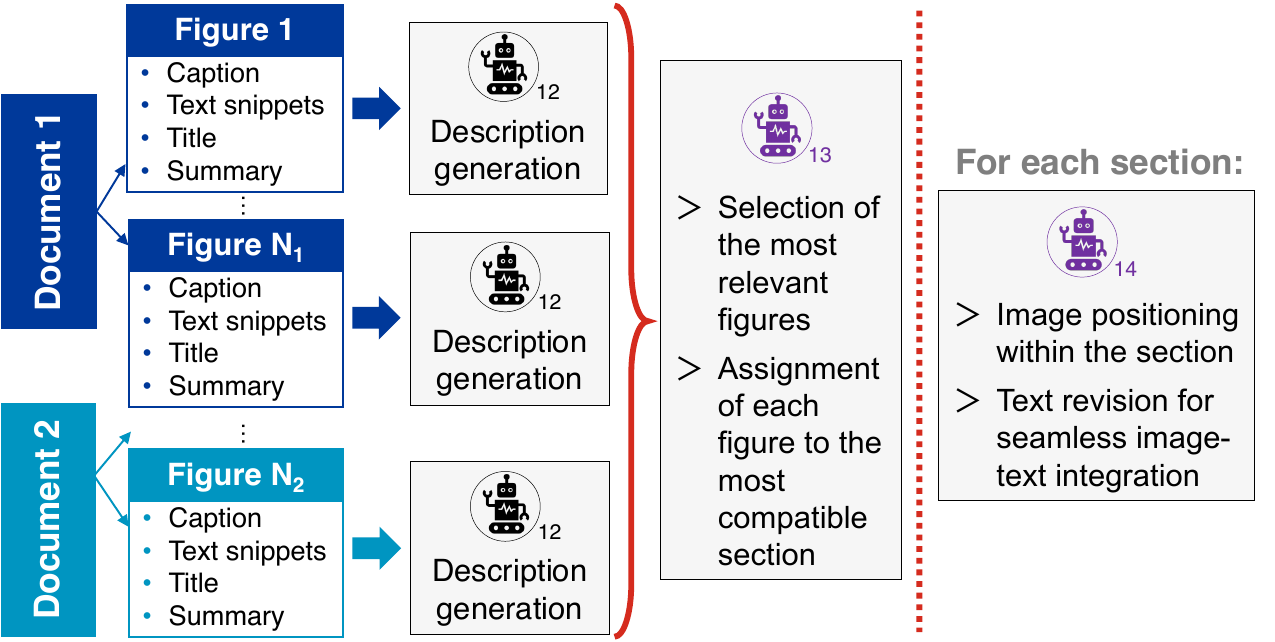}
  \caption{Structure of the figures selection and insertion routine. 
  The pipeline comprises the generation of the descriptions of all the images extracted from the sources, the selection of the most relevant figures, and their integration in the report.}
  \label{fig:method_images}
\end{figure}

Once the textual content of the report is generated, the workflow shifts to selecting, placing, and integrating figures within the report draft. The pipeline is illustrated in Figure~\ref{fig:method_images}. 

The pool of candidate images, denoted as $F$, comprises all the figures extracted from source documents. 
Agent~\emptycircled{12} is tasked with generating a description $d_f$ for each candidate figure $f \in F$. Only textual elements are used for the elaboration of the detailed depiction. Specifically, the agent retrieves the reference document title, its summary $r_s \in R_S$, the caption of the figure within the full-text document $r \in R$, and any in-text excerpt referring to the figure identifier extracted from the caption. 

Subsequently, reasoning agent~\emptycircled{13} evaluates all the figures' descriptions together, in conjunction with the report outline $O$, and selects the most relevant and informative images. If the length of the concatenation of the descriptions exceeds the context length of the \gls{llm}, this process is conducted hierarchically. The agent provides as output the set $\hat{F}$ of relevant figures, and assigns each of them to the most pertinent section of the report.

Then, reasoning agent~\emptycircled{14} carries out the placement of the figures, independently for each section. Given the set of figures $\hat{F}_s \subseteq \hat{F}$ assigned to section $s$, the agent has to assess their individual relevance for the specific section text, and images deemed as potentially out of scope are discarded. The key difference with respect to the selection performed a priori by agent~\emptycircled{13} is that the reasoning agent has access to the section text, thus has more elements to filter out images which do not convey useful information for the textual flow. Eventually, agent~\emptycircled{14} completes the images insertion by generating their captions and revising the section's text to integrate the figures seamlessly within the narrative.

\subsubsection{Overview table insertion}

To further enrich the report's content, Wyvern inserts an overview table summarizing the key concepts. First, reasoning agent~\emptycircled{15} analyzes the references set $R_S$, and extracts shared and informative aspects to differentiate among the relevant works. Based on this analysis, it builds a comparative table, along with a detailed explanation of its contents.
Subsequently, agent~\emptycircled{16} determines the most appropriate placement of the table within the outline~$O$, and outputs the name of the section in which the table should be integrated.
Finally, agent~\emptycircled{17} performs the actual table insertion. It receives the target section's text, the table and the accompanying explanation generated by agent~\emptycircled{15}. Its tasks are identifying the optimal insertion point within the specific section's text, generating a caption for the table, and editing the text of the section to ensure a meaningful content integration.

\subsubsection{Refinement}
Following the writing of the text and the inclusion of visual content, agent~\emptycircled{18} performs a final review to enhance the completeness and coherence of the report. In particular, it analyzes the content of each section, independently, to identify potential gaps or inconsistencies. When needed, it retrieves and inserts supplementary information from the references set $R_S$ to improve the depth and clarity of the section under consideration. When new content is inserted, the agent checks and possibly revises the citations. Additionally, it edits the text to ensure that the tone of the report remains neutral.

\subsection{Grounding}
\begin{figure}[]
  \centering
  \includegraphics[width=\linewidth]{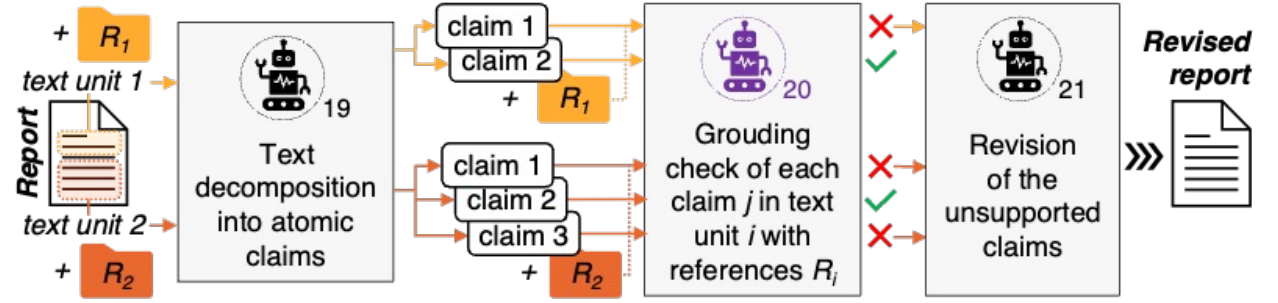}
  \caption{Overview of the grounding verification pipeline. The grounding of the claims is evaluated with their associated references, and then possibly revised.}
  \label{fig:method_grounding}
\end{figure}
Once the report has been generated, an ensemble of agents performs a final groundedness enhancement step. The objective of this phase is to review all the statements in the report and revise or remove them if they are not supported by the cited references.
This is achieved by first decomposing the text into atomic claims and then evaluating each of them independently, as illustrated in Figure~\ref{fig:method_grounding}.

More in detail, the text is segmented into a set $T$ of distinct text units, with paragraph-level granularity. For each text unit $t_i \in T$, the cited references constitute the references set $R_S^{i} \subseteq R_S$ for such textual element, which forms the basis for the subsequent grounding verification and content correction. Each text unit $t_i$ is then further decomposed by agent~\emptycircled{19} into a set $C_i$ of atomic textual claims, that all share the same reference set $R_S^i$. 

Following the decomposition, reasoning agent~\emptycircled{20} evaluates the grounding of each atomic claim $c_j \in C_i$ with respect to the support reference set, independently. First, it checks the claim against the set $R_S^i$, comprising the summaries of the associated references. If the claim is flagged as potentially not supported, the agent performs a second, more detailed entailment check with respect to the full-text of the references $R^i$, and the final \gls{nli} outcome is based on the second evaluation round. Each claim is also accompanied by a brief rationale that explains the agent's judgment.

Once all the atomic claims have been evaluated, agent~\emptycircled{21} carries out the revision routine. The agent operates on each text unit independently, but takes into account the broader context of the section to which the text unit belongs, in order to avoid redundancy or inconsistency during the refinement of the statements. Specifically, the agent considers the entire text of $t_i$'s section, the set of unsupported claims $C_i^{\text{unsupported}} \subseteq C_i$ to be modified within $t_i$ along with their accompanying grounding explanations, and the summaries of the references $R_S^i$. The agent then localizes each claim within the given text unit, and either revises it, in case it is only imprecise, or completely removes it.

Finally, agent~\emptycircled{22} operates on all the first-level sections independently, revising headings and subheadings to ensure consistency between sections' titles and content after the claims amendment.

\section{Experiments}
\label{sec:experiments}

\subsection{Experimental setup}
We implement our framework with LangChain.
For each input topic, we generate $n=30$ search queries. We use the Serper\footnote{\url{https://serper.dev/}} search API for the retrieval of the top-$k_1$ relevant resources, with $k_1=3$. For the embedded links extraction, we set $k_2=40$. We set the threshold values $\lambda_\text{comp}$ and $\lambda_\text{search}$ to 5, $\lambda_\text{link}$ to 7, and $\lambda_\text{sim}$ to 0.9. For the parsing of the PDF-based documents we use Docling~\cite{docling}, which allows also the extraction of figures, while we use the Playwright\footnote{\url{https://playwright.dev/}} Python library and Trafilatura~\cite{barbaresi-2021-trafilatura} for the other types of sources. We parse Wikipedia pages with the MediaWiki Action API\footnote{\url{https://www.mediawiki.org/wiki/API:Main\_page}} and the html2text\footnote{\url{https://pypi.org/project/html2text/}} library.
We employ DeepSeek-R1~\cite{deeseek-r1} to build all reasoning agents, and DeepSeek-V3~\cite{deepseek-v3} for all the other agents, both with temperature equal to 0.
We employ as evaluator models DeepSeek-V3 and Qwen3-32B~\cite{qwen3technicalreport}, with temperature of 1 and 0.6 respectively.

\subsection{Baselines}
We compare our method with STORM~\cite{shao2024storm}, WebThinker~\cite{Li2025WebThinker} and WikiAutoGen~\cite{yang2025wikiautogen}. 
For STORM, we use the Serper retriever with the default parameters as implemented by~\cite{shao2024storm}, and we replace the proprietary models with the open-source DeepSeek-V3. This substitution was made for reproducibility reasons, as proprietary models can be continuously updated over time, hindering the possibility of fair comparisons without a fixed snapshot of the models' weights.
For WebThinker, we employ the Serper retriever and the same open-source models as in~\cite{Li2025WebThinker}, i.e. WebThinker-QwQ-32B, a fine-tuned version of QwQ-32B, as reasoning model, and Qwen2.5-32B-Instruct as assistant model.
For WikiAutoGen, we use the Serper retriever and replace proprietary models with open-source alternatives, namely Pixtral Large~\cite{pixtral_large},  DeepSeek-R1 and DeepSeek-V3. 
For all the baselines, we use as input for the report generation the concatenation of the topic title and the detailed description of the aspects of interest.

\subsection{Human evaluation}
To evaluate the quality of the reports, we conduct a user study involving 27 volunteers from diverse corners of computing systems research, including senior researchers, engineers and Ph.D. students. To have evaluations of high technical quality, we ask each participant to select a topic of interest. We then provide each reviewer with two reports, one generated using Wyvern and one using a baseline method. We obtain nine distinct comparisons of Wyvern with respect to each baseline. We do not mention the used methodologies and we randomize the order between the two reports to ensure the absence of positional bias in the aggregated results.

We structure the questionnaire for the human evaluation in two main parts, i.e. the relative assessment and the absolute grading.
The relative assessment consists of a pairwise evaluation of two reports on various rubrics, concerning the structure, relevance, coverage, content presentation, figures and tables informativeness, engagement, and usefulness of the report. 
The main objective of the relative evaluation is to directly compare our method with the baselines. 
Absolute grading involves a more detailed evaluation on the same criteria, but applied to a single report, whose generation method is unknown to the participant. In particular, each participant assigns a score on a 5-point Likert scale~\cite{Likert1932A} to every statement in the questionnaire. This section of the evaluation enables a more precise assessment of the quality of the report generated with our proposed methodology.

\subsection{Automatic evaluation method}

\subsubsection{Grounding}
To measure the verifiability of the statements in the report we employ the citation recall and the citation precision, as in previous works~\cite{gao_enabling_2023, shao2024storm, wang2024autosurvey}. The citation recall $C_R$ measures the number of claims in the report that are supported by the cited references, i.e. given $N$ claims, where each claim $c_i$ is associated to a set of cited references $R_i$:
\begin{equation}
\label{eqn:citation_recall}
    \footnotesize
    C_R = \frac{\sum_{i=0}^{N-1}{f(c_i, R_i)}}{N}
\end{equation}
where $f(c_i, R_i)$ is the binary output of the \gls{nli} model for claim $i$, that assumes value 1 if the concatenation of the references $R_i$ supports the claim $c_i$, and 0 otherwise.

The citation precision measures the relevance of the citations in supporting the claims of the report. A citation $r_k \in R_i$ is considered relevant if: 
\begin{equation}
\footnotesize
    g(c_i,\!r_k) = \left[ \left[ f(c_i,\!r_k) \!=\! 1 \right]\!\lor\!\left[ f(c_i,\!R_i\!\setminus\!\{ r_k \})\!=\!0\right] \right]= 1
\end{equation}
\noindent where $[\cdot]$ denotes the Iverson bracket, which evaluates to 1 if the enclosed logical condition is true and 0 otherwise.
The overall citation precision, considering all the claims and the associated citations across all paragraphs in a document can be computed as:
\begin{equation}
\label{eqn:citation_precision}
\footnotesize
    C_P = \frac{\sum_{i=0}^{N-1} \sum_{k=0}^{|R_i|-1} f(c_i, R_i) \land g(c_i, r_k)}{\sum_{i=0}^{N-1} |R_i|}
\end{equation}

\subsubsection{Report quality}
We employ the evaluation rubrics proposed by~\citet{shao2024storm}, concerning the interest level of the report, its coherence and organization, relevance and focus, and broad coverage. We frame the evaluation as pairwise relative assessments, as in the human evaluation study. Namely, we provide as input to the evaluator models two reports, one generated by Wyvern, and the other by a baseline. We consider both order permutations, and we experimentally assess their effect on automated evaluation outcomes.

\section{Results}
\label{sec:results}
\subsection{Human evaluation}
\label{sec:results_human_evaluation}

\subsubsection{Pairwise relative comparisons}

\begin{figure}[]
  \centering
  \includegraphics[width=\linewidth]{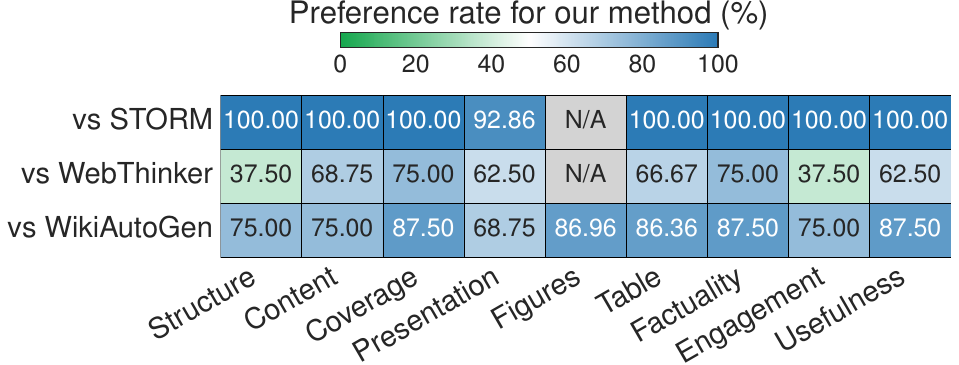}
  \caption{Relative grading of Wyvern with respect to selected baselines in the human evaluation study. Each cell reports the preference percentage for Wyvern over the baseline on a rubric. STORM and WebThinker cannot be compared on the \textit{Figures} criteria, as they produce text-only reports.}
\label{fig:results_human_relative}
\end{figure}

We report in Figure~\ref{fig:results_human_relative} the results of the evaluation study on the pairwise comparison between the reports, extracted from 23 completed questionnaires (out of the 27 distributed). Each cell contains the preference percentage of our method with respect to a considered baseline on a specific rubric. We cluster the various criteria that we ask the participants to evaluate into macro-areas, to provide a high-level overview. Additional details are given in the Appendices. Wyvern generates reports that surpass the ones produced by STORM on all the considered criteria. Additionally, it outperforms in 86.96\% of cases WikiAutoGen on the figures quality (informativeness, positioning, captioning). WebThinker achieves a higher preference rate (62.50\%) on the questions concerning the structure and the engagement level created by the report. While this highlights directions of improvement, Wyvern manages to produce reports that are qualitatively better on all the other considered aspects. In particular, users rate Wyvern's reports as more useful than STORM's, WebThinker's and WikiAutoGen's ones in 100.00\%, 62.50\% and 87.50\% of cases, respectively.

\subsubsection {Absolute grading}
\begin{figure}[]
  \centering
  \includegraphics[width=\linewidth]{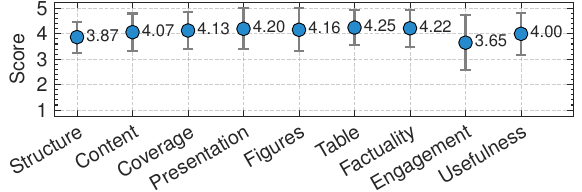}
  \caption{Results of the absolute grading of Wyvern's reports from the human evaluation study}
  \label{fig:results_human_absolute}
\end{figure}

To investigate the perceived quality of our framework, we include in the human evaluation study a finer-grained absolute scoring task of the reports generated by Wyvern.
Figure~\ref{fig:results_human_absolute} shows the results of this absolute grading, on the same rubrics considered in Figure~\ref{fig:results_human_relative}. Wyvern exhibits the highest scores on the rubrics concerning factuality and the overview table, with an absolute score of 4.22 and 4.25 out of 5, respectively. The engagement rubric receives the lowest score overall, equal to~3.65. However, it is also visible how the standard deviation of this criterion is the highest, highlighting the subjectivity of the question.

\subsection{Automatic evaluation}

\subsubsection{Grounding metrics}
\begin{figure}[]
  \centering
  \includegraphics[width=\linewidth]{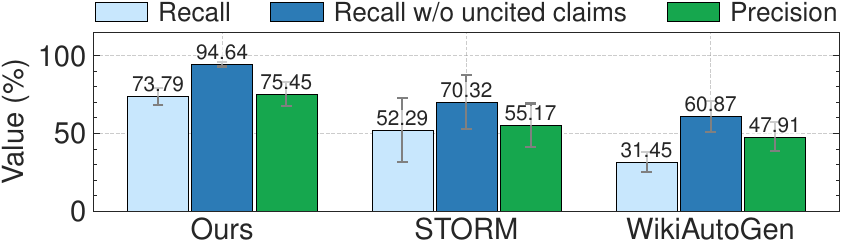}
  \caption{Citation quality metrics computed on all the reports generated by the various methods (27 documents in total for ours, 9 for STORM and WikiAutoGen). The error bar shows the standard deviation.}
  \label{fig:citation_metrics_results}
\end{figure}

We report in Figure~\ref{fig:citation_metrics_results} the citation recall and precision computed over the set of generated reports for each considered baseline. We were unable to compute such metrics for WebThinker, as it does not include citations in the report's text. Wyvern produces reports with higher grounding metrics than STORM and WikiAutoGen. In particular, it reaches 73.79\% of citation recall (+21.50 and +42.34 percentage points with respect to STORM and WikiAutoGen), and 94.64\% when considering only the claims with citations, thus excluding paragraphs without citations from the analysis. 

We further analyze the impact of the claims revision module on the grounding metrics computed over all the distributed reports in Table~\ref{tab:ablation_citation_metrics_autorevision}. Both citation recall and precision show a significant improvement after the application of this step. In particular, the citation recall increases from 60.92\% to 73.79\%, highlighting the positive effect of the claims revision stage on Wyvern's final output.

\begin{table}[]
\centering
\setlength{\tabcolsep}{2mm}
{\fontsize{9}{11}\selectfont
\begin{tabular}{cccc}
\hline
\textbf{Metric} &
  \textbf{Before (\%)} &
  \textbf{After (\%)} &
  \textbf{Gain} \\ \hline
\rowcolor[HTML]{EFEFEF} 
Cit. recall    & 60.92 ± 6.10  & 73.79 ± 5.44  & 1.21$\times$ \\
\begin{tabular}[c]{@{}c@{}}Cit. recall w/o \\ uncited claims\end{tabular} &
  84.34 ± 4.24 &
  94.64 ± 1.55 &
  1.12$\times$ \\
\rowcolor[HTML]{EFEFEF} 
Cit. precision & 67.15 ± 6.17 & 75.45 ± 7.54 & 1.12$\times$ \\ \hline
\end{tabular}%
}
\caption{Impact of the claims revision module on the citations quality metrics}
\label{tab:ablation_citation_metrics_autorevision}
\end{table}

\subsubsection{Pairwise relative comparisons}
\begin{figure}[]
  \centering
\includegraphics[width=\linewidth]{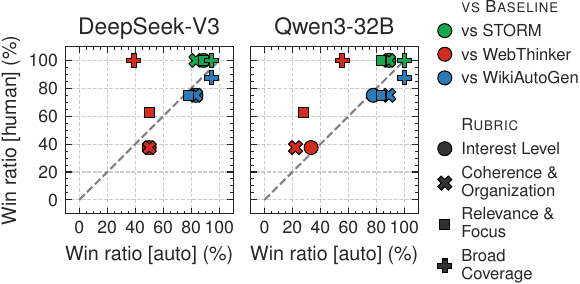}
  \caption{Comparison of human ($y$-axis) and automatic ($x$-axis) evaluation results on the four rubrics with two different evaluator models. The diagonal line represents a 1:1 correspondence between human and automatic evaluation.}
  \label{fig:results_auto_scatter}
\end{figure}

We perform an automatic evaluation, framed in a relative assessment scheme, adopting the four rubrics presented in~\cite{shao2024storm}, that cover the aspects of interest level, coherence and organization, relevance and focus, and broad coverage. We consider as evaluator models DeepSeek-V3 and Qwen3-32B. The goal is to compare the results obtained with the automatic evaluation with the ones of the human evaluation study, to assess whether the first can be used as proxy for the latter, being more scalable and cost- and time-effective.

Figure~\ref{fig:results_auto_scatter} reports the comparison between the results obtained over all the reports with different evaluator models, on the four criteria of judgment, and the human evaluation scores of the respective semantically equivalent question. 
The automatic means of evaluation (whose scores are reported on the $x$-axis) achieve a good correlation degree with respect to the human-assigned scores (on the $y$-axis)  for STORM and WikiAutoGen. However, they fail to capture the scores of WebThinker, especially on the \textit{Broad Coverage} and \textit{Relevance \& Focus} criteria. This highlights how \gls{llm}-based automatic evaluation of reports on qualitative rubrics still lacks behind in capturing the trends of the human-assigned evaluations, in particular when a clear preference gap is not always present in the latter (as it emerges in Figure~\ref{fig:results_human_relative}).

We extend our analysis by performing experiments to assess whether the evaluator models exhibit ordering bias, i.e. if the order in which the reports are passed as input to the model influences the final score. We consider two evaluator models, DeepSeek-V3 and Qwen3-32B. Ordering bias appears in 36.11\% and 30.56\% of the comparisons for DeepSeek-V3 and Qwen3-32B, respectively. This demonstrates that there is still an ample margin of improvement for achieving robustness in this type of automated evaluation.

\section{Conclusions}
\label{sec:conclusions}
We presented Wyvern, a multi-agent framework that enables the generation of grounded, multimodal technical reports on given input topics. 
We designed different modules, composed of ensembles of \gls{llm}-based agents, to handle the various stages of the report generation, including figures inclusion. Furthermore, with the integration of a final stage of text decomposition into atomic claims, verified on the collected evidence, Wyvern allows for an improvement in citation recall up to 2.3$\times$ with respect to the other considered methods, namely STORM, WebThinker, and WikiAutoGen. We conducted a human evaluation study to assess the quality of the reports generated by Wyvern, showing that their usefulness is perceived as higher in 63\% to 100\% of the instances, depending on the reference baseline. We additionally showed how the automatic means of evaluation lack accuracy and robustness in estimating human preference.

\section*{Limitations}
Wyvern relies on auxiliary search APIs to collect relevant documents from the web. Although our proposed framework includes specific sources filtering and selection pipelines, the generated multimodal report ultimately depends on the quality of the webpages retrieved by the search engine. To achieve broad coverage on the user-selected topic, Wyvern generates multiple, distinct search queries. However, this approach and the reliance on the search APIs do not always ensure a comprehensive and optimal coverage of the available information. Additionally, given the inherent characteristics of search APIs, reproducibility is not fully guaranteed, as search engines may return different results according to the inquiry time, location, and the always evolving indexing policies.

We evaluated Wyvern's performance exclusively on topics formulated in English language. Although a translator module could be inserted at the beginning of the pipeline to translate a topic expressed in any language to English (e.g., with an additional \gls{llm}-based agent), the generation of search queries in English language for the information retrieval penalizes (but not necessarily excludes) resources in other languages. The generalization capability of the framework to multilingual settings should be analysed in future works.

\section*{Ethical considerations}
Wyvern builds upon content retrieved from the web, which may inherently contain systematic biases as well as scientific inaccuracies. Furthermore, we rely on existing open-source \glspl{llm} that might not have been fully aligned with the ethical values of the reader. Currently, our framework does not incorporate an explicit mechanism for filtering or correcting biased or erroneous information. However, Wyvern does not amplify these issues beyond their presence in the original sources.

\section*{Acknowledgments}
We thank the evaluators who participated in the human evaluation study for providing their truthful judgment of the generated reports quality. 

This publication is part of the project PNRR-NGEU which has received funding from the MUR~–~DM 118/2023.

\bibliography{references}


\appendix

\section{Human evaluation}
We report in this section the complete, unaggregated list of questions and results of the relative assessment and absolute grading of the human evaluation study presented in the main text of the paper. All evaluators consented to the use of aggregated anonymous statistics collected from their responses, and of portions of the generated reports, for publication purposes.
\subsection{Relative assessment}
For the relative assessment in the human evaluation study, we provide each participant with two reports, named ``Report A" and ``Report B", without specifying the generation method. Each person receives one report generated by Wyvern, and one generated by another baseline. We divide the 27 participants into three groups of cardinality equal to 9, to have a uniform amount of comparisons over all the baselines. Furthermore, to avoid the presence of any positional bias in the participants' answers, we randomize the order and naming of the reports.
In Table~\ref{tab:table_human_relative_complete}, the list of questions and the associated answers from the human evaluation study are given.

Wyvern is systematically preferred over STORM on all the considered rubrics. WebThinker obtains a higher preference rate only on the structure, i.e. rubric (1), and on the engagement level of the report, namely rubric (15), where Wyvern reaches a 37.50\% favor rate. Regarding the structure of the report, the main criticalities that emerged concern the redundancy of the content over the sections and the overall length of the report. 
Wyvern and WebThinker achieve the same preference rate on rubric (6), which refers to the clarity of the explanations, and on rubric (4), related to the presentation of benefits and limitations about the topic. Wyvern excels on the topic coverage, i.e. rubric~(5), where it obtains a 100.00\% preference rate; and on the amount of technical details and factuality of the content (rubrics (3) and (15)), where it reaches 75.00\% of win percentage. When compared to WikiAutoGen, the only baseline that supports the inclusion of figures within the produced reports, Wyvern achieves preference rates of 85.71\%, 87.50\% and 87.50\% for rubrics (8), (9), and (10), which concern the quality of the positioning, the quality of the captions, and the informativeness degree of the figures, respectively. On all the other rubrics, Wyvern obtains a preference rate spanning from 62.50\% to 87.50\%.

\begin{table*}[]
\centering
\fontsize{9}{12}\selectfont
{\setlength{\tabcolsep}{2.9mm}
\begin{tabular}{clccc}
\hline
\multicolumn{1}{l}{} &
  \multicolumn{1}{c}{} &
  \multicolumn{3}{c}{\textit{\textbf{Preference rate for Wyvern}}} \\
\multicolumn{1}{l}{} &
  \multicolumn{1}{c}{\multirow{-2}{*}{\textbf{Question}}} &
  \multicolumn{1}{l}{\textbf{vs STORM}} &
  \textbf{vs WebThinker} &
  \textbf{vs WikiAutoGen} \\ \hline
\rowcolor[HTML]{EFEFEF} 
1 &
  Which report has the most logical structure? &
  100.00\% &
  37.50\% &
  75.00\% \\
2 &
  \begin{tabular}[c]{@{}l@{}}Which report better provides relevant information with\\ respect to the topic?\end{tabular} &
  100.00\% &
  62.50\% &
  75.00\% \\
\rowcolor[HTML]{EFEFEF} 
3 &
  \begin{tabular}[c]{@{}l@{}}Which report presents the best amount of technical \\ details about the topic?\end{tabular} &
  100.00\% &
  75.00\% &
  75.00\% \\
4 &
  \begin{tabular}[c]{@{}l@{}}Which report better presents both benefits/advantages \\ and challenges/limitations related to the topic?\end{tabular} &
  100.00\% &
  50.00\% &
  87.50\% \\
\rowcolor[HTML]{EFEFEF} 
5 &
  \begin{tabular}[c]{@{}l@{}}Which report has a broader coverage of multiple \\ aspects of the topic?\end{tabular} &
  100.00\% &
  100.00\% &
  87.50\% \\
6 &
  Which report has clearer explanations? &
  100.00\% &
  50.00\% &
  75.00\% \\
\rowcolor[HTML]{EFEFEF} 
7 &
  Which report has the most neutral tone? &
  85.71\% &
  75.00\% &
  62.50\% \\
8 &
  \begin{tabular}[c]{@{}l@{}}Which report has the most logical figures placement \\ (i.e., figures are assigned to a coherent section)? \\ If not applicable (e.g., there are no figures), \\ please select N/A.\end{tabular} &
  - &
  - &
  85.71\% \\
\rowcolor[HTML]{EFEFEF} 
9 &
  \begin{tabular}[c]{@{}l@{}}Which report has the most appropriate figures \\ descriptions and captions? If not applicable \\ (e.g., there are no figures), please select N/A.\end{tabular} &
  - &
  - &
  87.50\% \\
10 &
  \begin{tabular}[c]{@{}l@{}}Which report has the most informative figures? \\ If not applicable (e.g., there are no figures), \\ please select N/A.\end{tabular} &
  - &
  - &
  87.50\% \\
\rowcolor[HTML]{EFEFEF} 
11 &
  \begin{tabular}[c]{@{}l@{}}Which report has the most logical overview tables \\ placement (i.e., the overview table is assigned \\ to a coherent section)? If not applicable (e.g., there \\ is no overview table), please select N/A.\end{tabular} &
  \multicolumn{1}{l}{\cellcolor[HTML]{EFEFEF}100.00\%} &
  62.50\% &
  85.71\% \\
12 &
  \begin{tabular}[c]{@{}l@{}}Which report has the most appropriate overview \\ tables descriptions and captions? If not applicable \\ (e.g., there are no overview tables), please select N/A.\end{tabular} &
  \multicolumn{1}{l}{100.00\%} &
  75.00\% &
  85.71\% \\
\rowcolor[HTML]{EFEFEF} 
13 &
  \begin{tabular}[c]{@{}l@{}}Which report has the most informative overview \\ tables? If not applicable (e.g., there are no \\ overview tables), please select N/A.\end{tabular} &
  \multicolumn{1}{l}{\cellcolor[HTML]{EFEFEF}100.00\%} &
  62.50\% &
  87.50\% \\
14 &
  \begin{tabular}[c]{@{}l@{}}Which report has a better factuality to the best of \\ your knowledge?\end{tabular} &
  \multicolumn{1}{l}{100.00\%} &
  75.00\% &
  87.50\% \\
\rowcolor[HTML]{EFEFEF} 
15 &
  \begin{tabular}[c]{@{}l@{}}Which report is more engaging and \\ thought-provoking?\end{tabular} &
  \multicolumn{1}{l}{\cellcolor[HTML]{EFEFEF}100.00\%} &
  37.50\% &
  75.00\% \\
16 &
  Which report do you find the most useful? &
  \multicolumn{1}{l}{100.00\%} &
  62.50\% &
  87.50\% \\ \hline
\end{tabular}%
}
\caption{Results of the relative assessment section of the human evaluation study. Each percentage value represents the preference rate in the pairwise comparison between Wyvern and one of the baselines on a specific rubric. Comparisons cannot be made on criteria concerning figures when considering STORM and WebThinker as the other method, as explained in the paper.}
\label{tab:table_human_relative_complete}
\end{table*}
\begin{table*}[]
\centering
{
\fontsize{9}{12}\selectfont
\setlength{\tabcolsep}{2mm}
\begin{tabular}{cll}
\hline
\multicolumn{1}{l}{} &
  \multicolumn{1}{c}{\textbf{Question}} &
  \multicolumn{1}{c}{\textbf{Score}} \\ \hline
\rowcolor[HTML]{EFEFEF} 
1 &
  The structure of the report is logical. &
  \multicolumn{1}{c}{\cellcolor[HTML]{EFEFEF}3.87 ± 0.61} \\
2 &
  The report provides relevant information with respect to the topic. &
  \multicolumn{1}{c}{4.26 ± 0.53} \\
\rowcolor[HTML]{EFEFEF} 
3 &
  The report presents the right amount of technical details about the topic. &
  \multicolumn{1}{c}{\cellcolor[HTML]{EFEFEF}3.87 ± 0.85} \\
4 &
  The report presents both benefits/advantages and challenges/limitations related to the topic. &
  3.91 ± 0.78 \\
\rowcolor[HTML]{EFEFEF} 
5 &
  The report has a broad coverage of multiple aspects of the topic. &
  4.35 ± 0.56 \\
6 &
  The explanations of the report are: &
  3.91 ± 0.88 \\
\rowcolor[HTML]{EFEFEF} 
7 &
  The tone of the report is: &
  4.48 ± 0.65 \\
8 &
  \begin{tabular}[c]{@{}l@{}}The position of the figures within the report is logical (i.e., they are assigned to a coherent section).\\ If not applicable (e.g., there are no figures), please select N/A.\end{tabular} &
  4.52 ± 0.58 \\
\rowcolor[HTML]{EFEFEF} 
9 &
  \begin{tabular}[c]{@{}l@{}}The caption and the description of the figure's content in the text are appropriate. \\ If not applicable (e.g., there are no figures), please select N/A.\end{tabular} &
  3.87 ± 1.23 \\
10 &
  The figures are informative. If not applicable (e.g., there are no figures), please select N/A. &
  4.09 ± 0.72 \\
\rowcolor[HTML]{EFEFEF} 
11 &
  \begin{tabular}[c]{@{}l@{}}The position of the overview table within the report is logical (i.e., it is assigned to a coherent section). \\ If not applicable (e.g., there are no overview tables), please select N/A.\end{tabular} &
  4.50 ± 0.58 \\
12 &
  \begin{tabular}[c]{@{}l@{}}The caption and the description of the overview table's content in the text are appropriate. \\ If not applicable (e.g., there are no overview tables), please select N/A.\end{tabular} &
  4.14 ± 0.69 \\
\rowcolor[HTML]{EFEFEF} 
13 &
  \begin{tabular}[c]{@{}l@{}}The overview table is informative. If not applicable (e.g., there are no overview tables), \\ please select N/A.\end{tabular} &
  4.10 ± 0.75 \\
14 &
  To the best of your knowledge, the factuality of the report is: &
  4.22 ± 0.72 \\
\rowcolor[HTML]{EFEFEF} 
15 &
  How engaging and thought-provoking is the report? &
  3.65 ± 1.09 \\
16 &
  How useful do you find the generated report? &
  4.00 ± 0.83 \\ \hline
\end{tabular}%
}
\caption{Results of the absolute grading of the human evaluation study, reported with the mean and standard deviation of the given marks over the 23 questionnaires that were completed, out of the 27. Participants could assign a score in the [1,5]-range. For rubric (6), the interval extremes were associated with the adjectives ``confusing" and ``clear"; for rubric (7) with ``biased" and ``neutral"; and for rubric (14) with ``not satisfying" and ``very satisfying".}
\label{tab:results_human_absolute_complete}
\end{table*}
\subsection{Absolute grading}
We report in Table~\ref{tab:results_human_absolute_complete} the complete list of questions and results of the absolute grading section of the human evaluation study, before the clustering applied to obtain Figure~\ref{fig:results_human_absolute} of the paper. The considered rubrics are the same used for the relative assessment. As discussed in the paper, the absolute grading is used to have a more detailed overview of the quality of Wyvern's reports over distinct specific rubrics. The methodology of the report for which this evaluation was asked was unknown to the participants: only a generic report index (``Report A" or ``Report B") was given as indication of the target, hiding that it was corresponding to the one generated by Wyvern. For a performance comparison between Wyvern and the other baselines, refer to the relative assessment results.

From Table~\ref{tab:results_human_absolute_complete} it is possible to observe that the highest marks are obtained on rubrics (8) and (11), which concern the positioning of the figures and of the overview table in the report, with an average score of 4.52 and 4.50, respectively. The third highest average score, equal to 4.48, is achieved on rubric~(7), related to the neutrality of the tone, which is evaluated as satisfactorily neutral. The criterion that receives the lowest score, equal to 3.65, is the engagement and thought-provoking level of the technical report (rubric (15) in Table~\ref{tab:results_human_absolute_complete}). However, it can be seen that the standard deviation is quite high, which can be explained by the highly subjective nature of the answer. Also the structure of the report, namely rubric (1), and the textual integration of the figures within the text, i.e. rubric (12), exhibit margin of improvement, reaching a score of 3.87.
The factuality, the figures' informativeness and the usefulness of the report (rubrics (14), (10) and (16), respectively) are very positively evaluated, with average scores of 4.22, 4.09, and 4.00, demonstrating how Wyvern's design successfully accomplishes the objective of producing multimodal and grounded technical reports.

\section{Automatic evaluation}
We report in this section the complete results of the automatic evaluation of the reports, obtained considering the same four criteria presented in~\cite{shao2024storm}, which concern the interest level, the coherence and organization, the relevance and focus of the content, and the broad coverage of the topic. We consider both a relative assessment setup, discussed in the paper, and an absolute grading one.

\subsection{Relative assessment}
\begin{table}[]
\centering
\setlength{\tabcolsep}{1mm}
\resizebox{\columnwidth}{!}{
\begin{tabular}{cccc}
\hline
\textbf{Rubrics} & \textbf{vs STORM} & \textbf{vs WebThinker} & \textbf{vs WikiAutoGen} \\ \hline
\rowcolor[HTML]{EFEFEF} 
{\color[HTML]{000000} Interest level}                                & {\color[HTML]{000000} 50.00\%} & {\color[HTML]{000000} 50.00\%} & {\color[HTML]{000000} 50.00\%} \\
\begin{tabular}[c]{@{}c@{}}Coherence \& \\ Organization\end{tabular} & {\color[HTML]{000000} 50.00\%} & {\color[HTML]{000000} 50.00\%} & {\color[HTML]{000000} 50.00\%} \\
\rowcolor[HTML]{EFEFEF} 
\begin{tabular}[c]{@{}c@{}}Relevance \\ \& Focus\end{tabular}        & {\color[HTML]{000000} 50.00\%} & {\color[HTML]{000000} 50.00\%} & {\color[HTML]{000000} 50.00\%} \\
\begin{tabular}[c]{@{}c@{}}Broad \\ Coverage\end{tabular}            & {\color[HTML]{000000} 50.00\%} & {\color[HTML]{000000} 50.00\%} & {\color[HTML]{000000} 50.00\%} \\ \hline
\end{tabular}%
}
\caption{Results of the automatic relative assessment, performed using Prometheus~2-7B as evaluator model}
\label{tab:automatic_relative_results_prometheus}
\end{table}

In Figure~\ref{fig:results_auto_scatter} of the paper we presented the results of the pairwise relative evaluation, obtained using Qwen3-32B and DeepSeek-V3 as evaluator models. As discussed in the paper, for each pair of reports (one generated by Wyvern and the other by one of the baselines) we conduct the relative assessment twice, swapping the order of the reports, and assess the impact on the output of evaluators. As shown in Table~\ref{tab:ablation_citation_metrics_autorevision} of the paper, the two models did exhibit ordering bias, i.e. the outcome of the evaluation turned out to depend on the ordering of the reports under assessment.  
We carried out but omitted from the paper the same experiments using as evaluator Prometheus~2-7B~\cite{kim-etal-2024-prometheus2}, a model specifically trained on a relative ranking task. 

Table~\ref{tab:automatic_relative_results_prometheus} presents the results of this analysis, which shows that Prometheus~2-7B struggles even more to capture the human-assigned scores, giving the same preference rate to all the methods. By a manual inspection of the outcome, it was revealed that the evaluator always identifies as best report the first of the two options. Thus, since we consider both orderings when we perform the evaluation, the preference rate is always equal to 50.00\%, regardless of both method and criterion under consideration. 
The information position hence alters the final output of the model, as already shown in the literature~\cite{liu-etal-2024-lost}. 
We hypothesize that the cause of this behavior could lie in the fact that Prometheus~2-7B was trained with a limited maximum sequence length (4,096 tokens)~\cite{kim-etal-2024-prometheus2}, potentially affecting its capability to handle longer inputs properly.

\subsection{Absolute grading}
\begin{table*}[]
\centering
\fontsize{9}{12}\selectfont
{\setlength{\tabcolsep}{4mm}
\begin{tabular}{clcccc}
\hline
 &
  \multicolumn{1}{c}{} &
  \multicolumn{4}{c}{\textbf{Score}} \\
\multirow{-2}{*}{\textbf{Rubrics}} &
  \multicolumn{1}{c}{\multirow{-2}{*}{\textbf{Evaluator model}}} &
  \textbf{STORM} &
  \textbf{WebThinker} &
  \textbf{WikiAutoGen} &
  \textbf{Wyvern} \\ \hline
\rowcolor[HTML]{EFEFEF} 
\cellcolor[HTML]{EFEFEF}{\color[HTML]{000000} } &
  {\color[HTML]{000000} Prometheus-13B} &
  {\color[HTML]{000000} 4.00 ± 0.00} &
  {\color[HTML]{000000} \underline{ 4.56 ± 0.50}} &
  {\color[HTML]{000000} 4.00 ± 0.00} &
  {\color[HTML]{000000} 4.07 ± 0.47} \\
\rowcolor[HTML]{EFEFEF} 
\cellcolor[HTML]{EFEFEF}{\color[HTML]{000000} } &
  {\color[HTML]{000000} Prometheus~2-7B} &
  {\color[HTML]{000000} 4.22 ± 0.92} &
  {\color[HTML]{000000} 4.44 ± 0.83} &
  {\color[HTML]{000000} 4.67 ± 0.67} &
  {\color[HTML]{000000} \underline{4.85 ± 0.45}} \\
\rowcolor[HTML]{EFEFEF} 
\cellcolor[HTML]{EFEFEF}{\color[HTML]{000000} } &
  {\color[HTML]{000000} DeepSeek-V3} &
  {\color[HTML]{000000} 3.22 ± 0.63} &
  {\color[HTML]{000000} 4.44 ± 0.83} &
  {\color[HTML]{000000} 3.44 ± 0.68} &
  {\color[HTML]{000000} \underline{4.85 ± 0.52}} \\
\rowcolor[HTML]{EFEFEF} 
\multirow{-4}{*}{\cellcolor[HTML]{EFEFEF}{\color[HTML]{000000} \begin{tabular}[c]{@{}c@{}}Interest \\ Level\end{tabular}}} &
  {\color[HTML]{000000} Qwen3-32B} &
  {\color[HTML]{000000} 3.33 ± 0.47} &
  {\color[HTML]{000000} \underline{4.33 ± 0.67}} &
  {\color[HTML]{000000} 3.67 ± 0.67} &
  {\color[HTML]{000000} 3.78 ± 0.50} \\ \hline
{\color[HTML]{000000} } &
  {\color[HTML]{000000} Prometheus-13B} &
  {\color[HTML]{000000} \underline{4.78 ± 0.42}} &
  {\color[HTML]{000000} 4.44 ± 0.50} &
  {\color[HTML]{000000} 4.33 ± 0.47} &
  {\color[HTML]{000000} 3.96 ± 0.58} \\
{\color[HTML]{000000} } &
  {\color[HTML]{000000} Prometheus~2-7B} &
  {\color[HTML]{000000} 4.67 ± 0.47} &
  {\color[HTML]{000000} 4.67 ± 0.47} &
  {\color[HTML]{000000} 4.22 ± 0.42} &
  {\color[HTML]{000000} \underline{4.70 ± 0.46}} \\
{\color[HTML]{000000} } &
  {\color[HTML]{000000} DeepSeek-V3} &
  {\color[HTML]{000000} \underline{4.89 ± 0.31}} &
  {\color[HTML]{000000} 4.67 ± 0.47} &
  {\color[HTML]{000000} 4.22 ± 0.92} &
  {\color[HTML]{000000} 4.85 ± 0.36} \\
\multirow{-4}{*}{{\color[HTML]{000000} \begin{tabular}[c]{@{}c@{}}Coherence\\ \& \\ Organization\end{tabular}}} &
  {\color[HTML]{000000} Qwen3-32B} &
  {\color[HTML]{000000} 4.56 ± 0.68} &
  {\color[HTML]{000000} \underline{5.00 ± 0.00}} &
  {\color[HTML]{000000} 4.22 ± 0.63} &
  {\color[HTML]{000000} 4.52 ± 0.63} \\ \hline
\rowcolor[HTML]{EFEFEF} 
\cellcolor[HTML]{EFEFEF}{\color[HTML]{000000} } &
  {\color[HTML]{000000} Prometheus-13B} &
  {\color[HTML]{000000} \underline{4.89 ± 0.31}} &
  {\color[HTML]{000000} \underline{4.89 ± 0.31}} &
  {\color[HTML]{000000} 4.78 ± 0.63} &
  {\color[HTML]{000000} 4.11 ± 0.92} \\
\rowcolor[HTML]{EFEFEF} 
\cellcolor[HTML]{EFEFEF}{\color[HTML]{000000} } &
  {\color[HTML]{000000} Prometheus~2-7B} &
  {\color[HTML]{000000} \underline{5.00 ± 0.00}} &
  {\color[HTML]{000000} \underline{5.00 ± 0.00}} &
  {\color[HTML]{000000} \underline{5.00 ± 0.00}} &
  {\color[HTML]{000000} \underline{5.00 ± 0.00}} \\
\rowcolor[HTML]{EFEFEF} 
\cellcolor[HTML]{EFEFEF}{\color[HTML]{000000} } &
  {\color[HTML]{000000} DeepSeek-V3} &
  {\color[HTML]{000000} 4.78 ± 0.42} &
  {\color[HTML]{000000} 4.56 ± 0.50} &
  {\color[HTML]{000000} 4.00 ± 0.94} &
  {\color[HTML]{000000} \underline{4.81 ± 0.47}} \\
\rowcolor[HTML]{EFEFEF} 
\multirow{-4}{*}{\cellcolor[HTML]{EFEFEF}{\color[HTML]{000000} \begin{tabular}[c]{@{}c@{}}Relevance \\ \& \\ Focus\end{tabular}}} &
  {\color[HTML]{000000} Qwen3-32B} &
  {\color[HTML]{000000} 4.56 ± 0.50} &
  {\color[HTML]{000000} \underline{5.00 ± 0.00}} &
  {\color[HTML]{000000} 4.11 ± 0.31} &
  {\color[HTML]{000000} 4.59 ± 0.56} \\ \hline
{\color[HTML]{000000} } &
  {\color[HTML]{000000} Prometheus-13B} &
  {\color[HTML]{000000} \underline{5.00 ± 0.00}} &
  {\color[HTML]{000000} 4.89 ± 0.31} &
  {\color[HTML]{000000} \underline{5.00 ± 0.00}} &
  {\color[HTML]{000000} 4.19 ± 1.54} \\
{\color[HTML]{000000} } &
  {\color[HTML]{000000} Prometheus~2-7B} &
  {\color[HTML]{000000} \underline{4.44 ± 0.50}} &
  {\color[HTML]{000000} 4.22 ± 0.42} &
  {\color[HTML]{000000} 4.00 ± 0.00} &
  {\color[HTML]{000000} 4.41 ± 0.49} \\
{\color[HTML]{000000} } &
  {\color[HTML]{000000} DeepSeek-V3} &
  {\color[HTML]{000000} \underline{4.44 ± 0.50}} &
  {\color[HTML]{000000} 4.33 ± 0.47} &
  {\color[HTML]{000000} 4.00 ± 0.00} &
  {\color[HTML]{000000} 4.37 ± 0.48} \\
\multirow{-4}{*}{{\color[HTML]{000000} \begin{tabular}[c]{@{}c@{}}Broad \\ Coverage\end{tabular}}} &
  {\color[HTML]{000000} Qwen3-32B} &
  {\color[HTML]{000000} 4.44 ± 0.68} &
  {\color[HTML]{000000} 4.89 ± 0.31} &
  {\color[HTML]{000000} 4.67 ± 0.47} &
  {\color[HTML]{000000} \underline{4.93 ± 0.26}} \\ \hline
\end{tabular}%
}
\caption{Results of the automatic absolute grading, performed with different evaluator models, namely Prometheus-13B, Prometheus~2-7B, DeepSeek-V3, and Qwen3-32B. The results are reported in terms of the mean and standard deviation of the scores over all the reports generated by a given method. Underlined, the method achieving the highest average score with a specific evaluator on a given criterion.}
\label{tab:automatic_absolute_results_complete}
\end{table*}
We conduct an automatic evaluation in an absolute grading schema, similarly to~\cite{shao2024storm}. In these experiments, we provide only a single report to the evaluator model, which is prompted to assign a score in the [1,5]-range for each of the four considered criteria, i.e. \textit{Interest Level}, \textit{Coherence and Organization}, \textit{Relevance and Focus}, and \textit{Broad Coverage}. We employed as evaluator models Prometheus-13B~\cite{kim2024prometheus1}, Prometheus~2-7B~\cite{kim-etal-2024-prometheus2}, Qwen3-32B, and DeepSeek-V3. For Prometheus-13B, we employ the same iterative trimming strategy as in~\cite{shao2024storm} to reduce the length of text. However, we identify this evaluation setup as suboptimal, as for longer reports the majority of the content gets removed, impeding a fair evaluation. For this reason, we did not employ Prometheus-13B for the relative assessment, but only on the absolute grading with the same evaluation setup used in~\cite{shao2024storm}.
Table~\ref{tab:automatic_absolute_results_complete} presents the results of the experiments, that show that there is no uniform agreement between the evaluator models. For each rubric, distinct evaluators identify different optimal methods. This highlights how the choice of the evaluator model can highly affect the results and ranking between various methodologies, making it hard to obtain a fair and robust quantitative interpretation of the evaluations.

\subsection{Claims evaluation}
To verify whether automatic means of evaluation are effective for the grounding assessment, we conducted a manual evaluation comparing \glspl{llm} and human judgments on a sample of claims. Specifically, we randomly sampled 60 claims and verified whether the human annotations and the LLM judgments were consistent. The agreement rate was 93.33\%, indicating strong alignment between the two in this specific task.

\section{Cost analysis}

\begin{figure}[]
  \centering
  \includegraphics[width=\linewidth]{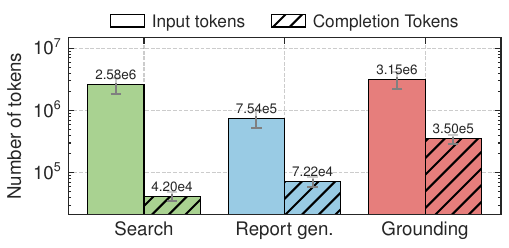}
  \caption{Input and completion tokens used by each module of Wyvern.}
  \label{fig:cost_per_module}
\end{figure}

\begin{figure}[]
  \centering
  \includegraphics[width=\linewidth]{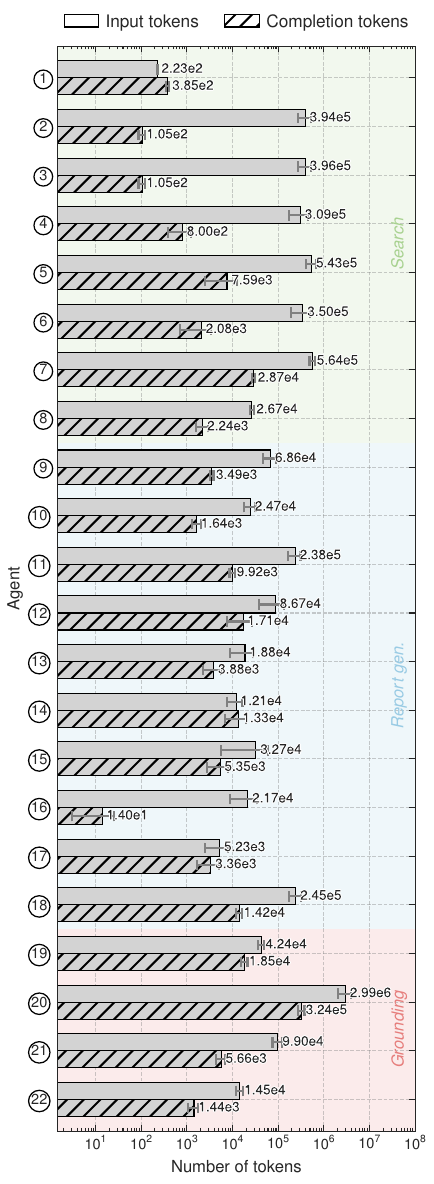}
  \caption{Input and completion tokens used by each agent of Wyvern.}
  \label{fig:cost_per_agent}
\end{figure}

We report the breakdown of the input and completion tokens of each module and agent of Wyvern in Figure~\ref{fig:cost_per_module} and Figure~\ref{fig:cost_per_agent}, respectively. Statistics are computed on a sample of 6 reports, using DeepSeek-V3 as base model and DeepSeek-R1-0528 as reasoning model. 
Agents \emptycircled{2}-\emptycircled{7} in the search module consume approximately the same number of input tokens, as they all analyze the collected documents individually.
Within the report generation module, agents \emptycircled{11} and \emptycircled{18} have high input and completion tokens usage as they are responsible for the report's text generation and refinement, based on the input references. Agent~\emptycircled{12} produces a significant amount of completion tokens due to the generation of image descriptions. 
The grounding module (especially agent \emptycircled{20}) required lengthy inputs mainly due to the two-stage grounding check, which may lead to considering many references (also in their full-text form); while the long outputs can be attributed to the long reasoning trace needed for the grounding assessment of each claim.

Regarding the parsing of the resources, which falls outside of the tokens usage analysis, we note that it can contribute significantly on the end-to-end latency of the framework, especially when a high number of long documents with many figures must be processed.

\section{Failure cases}
While the overall evaluation of our framework is positive, as detailed in Section~\ref{sec:results_human_evaluation}, the human evaluation study also served to identify the main weaknesses and directions of improvement. 

The main challenge noted by a small number of evaluators was the reports' excessive length and redundancy, sometimes resulting in a fragmented structure. A few evaluators also identified an uneven balance between breadth and depth, with too much focus given to peripheral topics.

Issues involving images, e.g. imprecise placement or explanation, instead stem mainly from insufficient textual descriptions in the input sources, which may lead to overly general figure descriptions by agent \emptycircled{12}. This in turn may lead to the placement of figures in sections not fully relevant to the figure's topic, or to overly vague in-text explanations of their content. Additionally, the parsing tool occasionally crops images imprecisely, resulting in minor inconsistencies between the images' content and their captions.

Regarding claims grounding, instances of only partially correct explanations and confusing terminology were also reported. 
Interestingly, we also came across an instance of a clearly false statement. After manually inspecting the cited reference, we found that the claim was indeed faithful to the source, and it was the source itself that contained the factual inaccuracy. As discussed in the \textit{Ethical considerations} section, we emphasize on this matter that while Wyvern targets the groundedness of the claims, it does not incorporate a mechanism to filter out erroneous information retrieved from the web.

\end{document}